\documentclass[10pt]{llncs}
\usepackage{times}
\usepackage{graphicx}
\usepackage{latexsym}
\usepackage{epic}
\usepackage{amsmath}
\usepackage{amssymb}
\usepackage{amsmath}

\usepackage{helvet}
\usepackage{courier}
\usepackage[noend]{algpseudocode}
\usepackage{algorithm}

\input{epsf}
\newcommand{\calC}{{\mathcal{C}}}
\newcommand{\calX}{{\mathcal{X}}}

\begin{document}
\newcommand{\mydiv}{\mbox{\rm div}}
\newcommand{\mymod}{\mbox{\scriptsize \rm \ mod \ }}
\newcommand{\mymin}{\mbox{\rm min}}
\newcommand{\mymax}{\mbox{\rm max}}

\newcommand{\calA}{{\cal A}}
\newcommand{\calL}{{\cal L}}
\newcommand{\dom}{{dom}}
\newcommand{\real}{\ensuremath{\mathbb{R}}}
\newcommand{\nat}{\ensuremath{\mathbb{N}}}
\newcommand{\ZZ}{\ensuremath{\mathbb{Z}}}

\newcommand{\set}{\mathcal}
\newcommand{\myset}[1]{\ensuremath{\mathcal #1}}
\newcommand{\myomit}[1]{}
\newcommand{\tighter}{\mbox{$\preceq$}}
\newcommand{\stighter}{\mbox{$\prec$}}
\newcommand{\incomparable}{\mbox{$\bowtie$}}
\newcommand{\equivalent}{\mbox{$\equiv$}}

\newcommand{\reg}{\mbox{$RE$}}

\newcommand{\gsc}{\mbox{\sc Gsc}}
\newcommand{\gcc}{\mbox{\sc Gcc}}
\newcommand{\GCC}{\mbox{\sc Gcc}}
\newcommand{\AllDifferent}{\mbox{\sc AllDifferent}}
\newcommand{\ALLDIFF}{\mbox{\sc AllDifferent}}


\newcommand{\SLIDE}{\mbox{\sc Slide}}
\newcommand{\SLIDINGSUM}{\mbox{\sc SlidingSum}}
\newcommand{\TABLE}{\mbox{\sc Table}}
\newcommand{\CIRCREGULAR}{\mbox{$\mbox{\sc Regular}_{\odot}$}}
\newcommand{\WRAPREGULAR}{\mbox{$\mbox{\sc Regular}^{*}$}}
\newcommand{\STRETCH}{\mbox{\sc Stretch}}
\newcommand{\INCSEQ}{\mbox{\sc IncreasingSeq}}
\newcommand{\INC}{\mbox{\sc Increasing}}
\newcommand{\lseX}{\mbox{\sc Lex}}
\newcommand{\NFA}{\mbox{\sc NFA}}
\newcommand{\DFA}{\mbox{\sc DFA}}

\newcommand{\PRECEDENCE}{\mbox{\sc Precedence}}
\newcommand{\lseXVAR}{\mbox{\sc LexLeader}}
\newcommand{\lseXGENSET}{\mbox{\sc SetGenLexLeader}}
\newcommand{\lseXSETVAR}{\mbox{\sc SetLexLeader}}
\newcommand{\lseXMSETVAR}{\mbox{\sc MSetLexLeader}}
\newcommand{\lseXSETVAL}{\mbox{\sc SetValLexLeader}}
\newcommand{\lseXVAL}{\mbox{\sc ValLexLeader}}
\newcommand{\lseXVALVAR}{\mbox{\sc GenLexLeader}}
\newcommand{\VALVARLEX}{\mbox{\sc ValVarLexLeader}}
\newcommand{\NVALUES}{\mbox{\sc NValues}}
\newcommand{\USES}{\mbox{\sc Uses}}
\newcommand{\COMMONG}{\mbox{\sc Common}}
\newcommand{\CARDPATH}{\mbox{\sc CardPath}}
\newcommand{\RANGE}{\mbox{\sc Range}}
\newcommand{\ROOTS}{\mbox{\sc Roots}}
\newcommand{\AMONG}{\mbox{\sc Among}}
\newcommand{\ATMOST}{\mbox{\sc AtMost}}
\newcommand{\ATLEAST}{\mbox{\sc AtLeast}}
\newcommand{\ATMOSTSEQ}{\mbox{\sc AtMostSeq}}
\newcommand{\ATLEASTSEQ}{\mbox{\sc AtLeastSeq}}
\newcommand{\AMONGSEQ}{\mbox{\sc AmongSeq}}
\newcommand{\SEQUENCE}{\mbox{\sc Sequence}}
\newcommand{\GENSEQUENCE}{\mbox{\sc Gen-Sequence}}
\newcommand{\SEQ}{\mbox{\sc Seq}}
\newcommand{\myelement}{\mbox{\sc Element}}
\newcommand{\LEX} {\mbox{\sc Lex}}
\newcommand{\REPEAT} {\mbox{\sc Repeat}}
\newcommand{\REPEATONE} {\mbox{\sc RepeatOne}}
\newcommand{\STRETCHREPEAT} {\mbox{\sc StretchRepeat}}
\newcommand{\STRETCHONEREPEAT} {\mbox{\sc StretchOneRepeat}}
\newcommand{\STRETCHONEREPEATONE} {\mbox{\sc StretchOneRepeatOne}}
\newcommand{\SETSIGLEX} {\mbox{\sc SetSigLex}}
\newcommand{\SETPREC} {\mbox{\sc SetPrecedence}}

\newcommand{\SOFTATMOSTSEC} {\mbox{\sc SoftAtMostSequence}}
\newcommand{\ATMOSTSEC} {\mbox{\sc AtMostSequence}}
\newcommand{\SOFTATMOST} {\mbox{\sc SoftAtMost}}
\newcommand{\SOFTSEQ} {\mbox{\sc SoftSequence}}
\newcommand{\SOFTAMONG} {\mbox{\sc SoftAmong}}
\newcommand{\SEQCYC}{\mbox{$\mbox{\sc CyclicSequence}$}}

\newcommand{\ATMOSTSEQCYC}{\mbox{$\mbox{\sc AtMostSeq}_{\odot}$}}
\newcommand{\ignore}[1]{}
\newtheorem{assertion}{Assertion}
\newcommand{\mypair}[2]{\mbox{$\langle \mbox{\rm #1},\mbox{\rm #2} ]$}}

\newcommand{\ninaS}[1]{{#1}}
\newcommand{\nina}[1]{{#1}}


\newcommand{\mreg}{\mbox{$MR$}}
\newcommand{\hprs}{\mbox{$HPRS$}}
\newcommand{\reglo}{\mbox{$LO$}}
\newcommand{\ps}{\mbox{$PS$}}
\newcommand{\tc}{\mbox{$ST$}}
\newcommand{\among}{\mbox{$AD$}}
\newcommand{\lse}{\mbox{$LG$}}
\newcommand{\lser}{\mbox{$LG_R$}}
\newcommand{\cs}{\mbox{$CS$}}
\newcommand{\csdc}{\mbox{$CS_{DC}$}}
\newcommand{\fl}{\mbox{\sc $FB$}}

\newcommand{\REGULAR}{\mbox{\sc Regular}}

\newtheorem{mytheorem}{Theorem}
\newcommand{\myproof}{\noindent {\bf Proof:\ \ }}
\newcommand{\myqed}{\mbox{$\diamond$}}
\newcommand{\LEXCH}{\mbox{\sc LexChain}}
\newcommand{\BETWEEN}{\mbox{\sc Between}}
\newcommand{\COMPLB}{\mbox{\sc CompLB}}
\newcommand{\COMPUB}{\mbox{\sc CompUB}}
\newcommand{\CHAIN}{\mbox{\sc Chain}}
\newcommand{\MATRIXLEX}{\mbox{\sc MatrixLex}}
\newcommand{\SEQUENCEBT}{\mbox{\sc SequenceBt}}
\newcommand{\REGULARBT}{\mbox{\sc RegularBt}}
\newcommand{\GCCBT}{\mbox{\sc GCCBt}}
\newcommand{\CBT}{\mbox{\sc Cbt}}
\newcommand{\CGR}{\mbox{\sc Cgrt}}
\newcommand{\SUM}{\mbox{\sc Sum}}
\newcommand{\LEXX}{\mbox{\sc C\&Lex}}
\newcommand{\LEXSOL}{\mbox{\sc LexSol}}
\newcommand{\COSTGCC}{\mbox{\sc CostGCC}}

{\makeatletter
 \gdef\xxxmark{%
   \expandafter\ifx\csname @mpargs\endcsname\relax 
     \expandafter\ifx\csname @captype\endcsname\relax 
       \marginpar{xxx}
     \else
       xxx 
     \fi
   \else
     xxx 
   \fi}
 \gdef\xxx{\@ifnextchar[\xxx@lab\xxx@nolab}
 \long\gdef\xxx@lab[#1]#2{{\bf [\xxxmark #2 ---{\sc #1}]}}
 \long\gdef\xxx@nolab#1{{\bf [\xxxmark #1]}}
}

\newcommand{\new}[1]{{#1}}
\newcommand{\myvec}[1]{\mbox{\boldmath $#1$}}

\pagestyle{plain}

\title{Combining Symmetry Breaking and Global
Constraints\thanks{NICTA is funded by 
the Australian Government as represented by 
the Department of Broadband, Communications and the Digital Economy and 
the Australian Research Council through the ICT Centre of Excellence program. 
}}

\author{George Katsirelos \and
Nina Narodytska \and
Toby Walsh}
\institute{NICTA and UNSW, Sydney, Australia
}
\maketitle

\begin{abstract}
We propose a new family of constraints
which combine together lexicographical ordering
constraints for symmetry breaking with 
other common global constraints. We give
a general purpose propagator for this family
of constraints, and show how to improve its 
complexity by exploiting properties of the included
global constraints. 
\end{abstract}
\sloppy
\section{Introduction}

The way that a problem is modeled is critically important
to the success of constraint programming. 
Two important aspects of modeling
are symmetry and global constraints.
A common and effective method of dealing
with symmetry is to introduce constraints
which eliminate some or all of the symmetric
solutions \cite{puget93}. Such symmetry breaking 
constraints are usually considered separately to other
(global) constraints in a problem. However,
the interaction between problem and symmetry breaking constraints
can often have a significant impact on search. For 
instance, the interaction between problem and symmetry breaking
constraints gives an exponential reduction in
the  search required to solve certain pigeonhole problems
\cite{Walsh07Sym2}. In this paper, we consider even tighter links
between problem and symmetry breaking constraints.
We introduce a family of global constraints
which combine together a common type of symmetry breaking
constraint with a range of common problem constraints.
This family of global constraints is
useful for modeling scheduling, rostering and 
other problems. 

Our focus here is on matrix models \cite{matrix2002}. Matrix models
are constraint programs containing matrices of 
decision variables on which common patterns of constraints
are posted. For example, in a 
rostering problem, we might have a matrix of
decision variables with the rows representing
different employees and the columns representing 
different shifts. A problem constraint
might be posted along each row to ensure
no one works too many night shifts in any 7 day period, 
and along each column to ensure sufficient employees work
each shift. A common type of symmetry on such matrix models
is row interchangeability  \cite{ffhkmpwcp2002}.
Returning to our rostering example,
rows representing equally skilled 
employees might be interchangeable.
An effective method to break such symmetry is
to order lexicographically the rows of the matrix\cite{ffhkmpwcp2002}. 
To increase the propagation between such 
symmetry breaking and problem constraints, 
we consider compositions of 
lexicographical ordering and problem constraints. 
We conjecture that the additional pruning achieved
by combining together symmetry breaking and 
problem constraints will justify the additional
cost of propagation. In support of this, we 
present a simple problem where it gives a super-polynomial
reduction in search. 
We also implement these new propagators
and run them on benchmark nurse scheduling problems. 
\ninaS{Experimental results show that propagating of a combination  of
symmetry breaking and global constraints reduces the search space significantly and improves run time for most of the benchmarks.}

\section{Background}

A constraint satisfaction problem (CSP) $P$ consists of a set of
variables $\calX = \{X[i]\}$, $i=1,\ldots,n$ each of which has a finite domain $D(X[i])$, and a
set of constraints $\calC$. We use capital letters for variables (e.g.
$X[i]$ or $Y[i]$), lower case for values (e.g. $v$ or $v_i$)
and write $\myvec{X}$ for the sequence of variables,
$X[1]$ to $X[n]$.
A constraint $C \in \calC$ has a
\emph{scope}, denoted $scope(C) \subseteq \calX$ and allows a subset of
the possible assignments to the variables $scope(C)$, 
called \emph{solutions} or \textit{supports} of $C$. 
A constraint is \emph{domain consistent} (\emph{DC}) iff for each
variable $X[i]$, 
every value in the domain of $X_i$ belongs to a support.
A solution of a CSP $P$ is an assignment of 
one value to each variable such that
all constraints are satisfied.
A matrix model of a CSP is one in which
there is one (or more) matrices of decision variables.
For instance, in a rostering problem, one dimension might represent 
different employees and the other dimension might represent
days of the week. 


A common way to solve a CSP is with backtracking search. In
each node of the search tree, a \emph{decision} restricts the domain
of a variable 
and the 
solver infers
the effects of 
that decision by invoking a \emph{propagator} for each constraint. 
A propagator for a constraint $C$ is an algorithm which
takes as input the domains of the variables in $scope(C)$ and returns
\emph{restrictions} of these domains. 
We say the a propagator enforces \emph{domain consistency (DC)} on a constraint $C$
iff an invocation of the propagator ensures that the constraint $C$ is domain consistent.
%

A \emph{global constraint} is a constraint in which the number of variables
is not fixed. Many common and useful global constraints have been proposed.
We introduce here the global constraints used in this paper.
The global lexicographical ordering constraint
$\LEX(\myvec{X},\myvec{Y})$ 
is recursively defined to hold
iff $X[1] < Y[1]$, or 
$X[1]=Y[1]$ and $\LEX([X[2],\ldots,X[n]],[Y[2],\ldots,Y[n]])$
\cite{Frisch02}. This constraint is used to break symmetries between vectors
of variables.
The global sequence constraint
$\SEQUENCE(l,u,k,\myvec{X},V)$ holds
iff $l \leq |\{ i \ | \ X[i]\in V, j \leq i < j+k\}| \leq u$ for each 
$1 \leq j < n-k$ \cite{Beldiceanu94CHIP}. 
The regular language constraint
$\REGULAR(\mbox{$\cal A$},\myvec{X})$
holds iff $X[1]$ to $X[n]$ 
takes a sequence of values accepted by 
the deterministic finite automaton 
\mbox{$\cal A$} \cite{Pesant04}.
The last two constraints are useful in  modeling rostering
and scheduling problems.

\section {The $\LEXX$ constraint}
\label{s:gen_lexx}

Two common patterns in many matrix models are that
rows of the matrix are interchangeable,
and that a global constraint $C$ is applied to each row. 
To break such row symmetry, we can post 
constraints that lexicographically order rows \cite{ffhkmpwcp2002}. 
To improve propagation between 
the symmetry breaking and problem constraints,
we propose the $\LEXX(\myvec{X},\myvec{Y},C)$ constraint. 
This holds iff $C(\myvec{X})$, $C(\myvec{Y})$
and $\LEX(\myvec{X},\myvec{Y})$ all simultaneously
hold. 
To illustrate the potential value of such a \LEXX\ constraint,
we give a simple example where it reduces 
search super-polynomially.

\begin{example}
\label{e:sep}
Let $M$ be a $n \times 3$ matrix in which all rows are interchangeable.
Suppose that $C(X,Y,Z)$ 
ensures $Y = X + Z$, and
that variable domains are as follows:

$M=\left(
\begin{matrix}	
  \{ 1,\ldots, n-1 \} & \{ n+1,\ldots,2n-1 \} & n\\
  \{ 1,\ldots, n-1 \} & \{ n,\ldots,2n-2 \} & n-1\\
   \ldots  &  \ldots & \ldots\\
  \{1,\ldots, n-1\} & \{3,\ldots,n+1\} & 2\\
  \{1,\ldots, n-1\} & \{2,\ldots,n\} & 1
\end{matrix}  
\right)$.

We assume that the branching heuristic
instantiates variables top down and left to right,
trying the minimum value first.
We also assume we enforce DC on posted constraints.
If we model the problem with \LEXX\ constraints, 
we solve it without search.
On the other hand, if we model the problem 
with separate \LEX\ and $C$ constraints, 
we explore an exponential sized search tree before 
detecting inconsistency using the mentioned branching
heuristic and a super-polynomial sized tree with any $k$-way branching 
heuristic. 

\myomit{
We give outline of the proof here.
Firstly, we will show that posting $\LEXX(M^{<i>},M^{<i+1>})$, $i=1,\ldots,n-1$ 
on each row  detects the inconsistency in the root node. 
The $\LEXX(M^{<1>},M^{<2>})$ constraint removes
the value $\{1\}$ from the domain of $X[2]$, the $\LEXX(M^{<2>},M^{<3>})$ constraint removes
values $\{1,2\}$ from the domain of $X[3]$, and so on. Finally,  
$\LEXX(M^{<n-1>},M^{<n>})$ prunes values $\{1,\ldots,n-1\}$ from
$X[n]$.

Secondly, we show that the decomposition of $\LEXX(M^{<i>},M^{<i+1>})$, $i=1,\ldots,n-1$ into
individual constraints explores an exponential space to  detect 
inconsistency. We notice a special property of this problem. If we do a shift transformation 
for variables $\myvec{X}$ and $\myvec{Y}$, namely $X'[i]= X[i] + s$ and $Y'[i]= Y[i] + s$, $i=1,\ldots, n$, 
where $s$ is a shift, we obtain an equivalent problem. After 
the transformation, constraints $C(X'Y'Z[i])$ ensure that $Y'[i] = X'[i] + Z[i]$, $i=1,\ldots, n$. 
Hence any solution of the transformed problem can be 
converted to a solution of the original problem by subtracting $s$ from $X$ and $Y$.
This observation is used to construct a recursive formula for the search tree size.
Let $P(n,s)$ be the search tree of the problem with $n$ rows in matrix $M$
and a shift $s$. Let $S(n)$ be the size of the tree for $P(n,s)$, $s=[-\inf,+\inf]$.
Note that the size of the tree does not depend on the shift.
It can be proved by induction on the number of rows that the size
of the tree grows as $S(n) = 2S(n-1)+1$ which is exponential in the size of the problem.

}

\end{example}

\subsection{Propagating \LEXX}
\label{ss:xlex}
We now show how, given a (polynomial time) 
propagator for the constraint $C$, we can
build a (polynomial time) propagator
for \LEXX . 
The propagator \ninaS{is inspired by the $DC$ filtering algorithm for the $\LEXCH$ constraint
proposed by Carlsson and Beldiceanu~\cite{Carlsson02LEXCHAIN}.
The $\LEXCH$ constraint ensures that rows of the matrix $M$ are lexicographically
ordered. If the $\LEXCH$ constraint 
is posted on two rows then $\LEXCH$ is equivalent to the $\LEXX(X,Y,\mathrm{True})$ constraint.
However, unlike ~\cite{Carlsson02LEXCHAIN},  we can propagate here
a conjunction of the $\LEX$ constraint and arbitrary global constraints $C$}. 
The propagator for the $\LEXX$ constraint is based on the following result 
which decomposes propagation into two simpler problems.

\sloppy  
\begin {proposition}
\label{p:p1}
Let $\myvec{X_l}$ be the lexicographically smallest solution of $C(\myvec{X})$,
$\myvec{Y_u}$ be the lexicographically greatest 
solution of $C(\myvec{Y})$, and $\LEX(\myvec{X_l},\myvec{Y_u})$.
Then enforcing $DC$ on $\LEXX(\myvec{X},\myvec{Y},C)$
is equivalent to enforcing $DC$ on 
$\LEXX(\myvec{X}, \myvec{Y_u},C)$ and on $\LEXX(\myvec{X_l},\myvec{Y},C)$ .  
\end {proposition}
\proof
Suppose $\LEXX(\myvec{X_l},\myvec{Y},C)$ is DC. 
We are looking for support for $Y_k=v$, where $Y_k$ is an arbitrary variable in $\myvec{Y}$.
Let $\myvec{Y'}$ be a support for $Y_k=v$ in $\LEXX(\myvec{X_l},\myvec{Y},C)$.
Such a support exists because  $\LEXX(\myvec{X_l},\myvec{Y},C)$ is DC.
$\LEXX(\myvec{X_l},\myvec{Y},C)$ ensures that $\myvec{Y'}$ is a solution 
of $C(\myvec{Y})$ and $\LEX(\myvec{X_l},\myvec{Y'})$.
Consequently, $\myvec{X_l}$ and $\myvec{Y'}$ are a solution 
of $\LEXX(\myvec{X},\myvec{Y},C)$. 
Similarly, we can find a support for $X_k=v$, where $X_k$ is an arbitrary variable in $\myvec{X}$.
\qed

Thus, we will build a propagator for \LEXX\ 
that constructs the lexicographically smallest (greatest) 
solution of $C(\myvec{X})$ ($C(\myvec{Y})$) 
and then uses two simplified $\LEXX$
constraints in which the first (second) sequence of variables is 
replaced by the appropriate bound. 

\subsubsection{\label{ss:findlexmin} Finding the lexicographically smallest solution.}

We first show how to find the lexicographically smallest solution 
of a constraint. \ninaS{We denote this algorithm   $C_{min}(\myvec{L},\myvec{X})$}. 
A dual method is used to find
the lexicographically greatest solution. We use a greedy 
algorithm that scans through $\myvec{X}$ 
and extends the partial solution by selecting 
the smallest value from the domain of $X[i]$ at $i$th step 
(line \ref{a:s_c_min_min}).
To ensure that the selection at the next step will never 
lead to a failure, the algorithm enforces $DC$ after each value 
selection (line \ref {a:s_c_min_dc}). Algorithm~\ref{a:c_min} 
gives the pseudo-code for the $C_{min}(\myvec{L},\myvec{X})$ algorithm.
The time complexity of Algorithm~\ref{a:c_min} is $O(nc + nd)$,
where $d$ is the total number of values in the domains of variables $\myvec{X}$ and
$c$ is the (polynomial) cost of enforcing DC on $C$.

\begin{algorithm}
\scriptsize {
\caption{$C_{min}(\myvec{L},\myvec{X})$ }\label{a:c_min}
\begin{algorithmic}[1]
\Procedure{$C_{min}$}{$\myvec{L}:out,\myvec{X}:in$}
\If {$(DC(C(\myvec{X})) == fail)$}
	\State return $false$; \label{a:c_min_fail} 
\EndIf
\State $\myvec{Y} = Copy(\myvec{X})$;	
\For{$i = 1$ \textbf{to} $n$} \label{a:s_c_min}								
	\State $Y[i] = L[i] = \min (D(Y[i]))$; \label{a:s_c_min_min}
	\State $DC(C(\myvec{Y}))$;\label{a:s_c_min_dc}
\EndFor \label{a:e_c_min}			

\State return $true$;
\EndProcedure
\end{algorithmic}
}
\end{algorithm}

\begin {proposition}
\label{p:al_1_proof}
\nina{ Let $C(\myvec{X})$ be a global constraint.
Algorithm~\ref{a:c_min} returns the lexicographically smallest solution of the global constraint $C$ if such
a solution exists.}
\end {proposition}
\proof
First we prove that if there is a solution 
to $C(\myvec{X})$ then Algorithm~\ref{a:c_min}
returns a solution. Second, 
we prove that the solution returned is the lexicographically smallest
solution. 
\begin{enumerate}
	\item If $C(\myvec{X})$ does not have a solution then Algorithm~\ref{a:c_min} fails at line \ref{a:c_min_fail}.
	 Otherwise $C(\myvec{X})$ has a solution.  Since $DC(C(\myvec{X}))$ leaves only consistent values,  any value of $X[1]$ can be extended to a solution of $C(\myvec{X})$ and Algorithm~\ref{a:c_min} selects 
	 $L[1]$ to be the minimum value of $X[1]$. Suppose Algorithm~\ref{a:c_min} performed $i-1$ steps and the partial
	 solution is $[L[1], \ldots, L[i-1]]$. All values left in the domains of at $X[i],\ldots,X[n]$ 
	 are consistent with  the partial solution $[L[1], \ldots, L[i-1]]$. Consequently, any value that is in the domain of $X[i]$ is 
	 consistent with $[L[1], \ldots, L[i-1]]$ and can be extended to a solution of $C(\myvec{X})$. The algorithm assigns $L[i]$ to the minimum
	 value of $X[i]$. Moving forward to the end of the sequence, the algorithm finds a solution to $C(\myvec{X})$.
	\item By contradiction. Let $\myvec{L'}$ be the lexicographically smallest solution of  $C(\myvec{X})$ and $\myvec{L}$ be the solution returned by 
	Algorithm~\ref{a:c_min}. Let $i$ be the first position where $\myvec{L'}$ and $\myvec{L}$ differ so  that $L'[i] < L[i]$,
	$L'[k] = L[k]$, $k=1,\ldots, i-1$. Consider $i$th step of Algorithm~\ref{a:c_min}. As $DC(C(\myvec{X}))$ is correct, 
	all values of $X[i]$ consistent with $[L[1],\ldots, L[i-1]]$ are in the domain of $X[i]$. The algorithm selects $L[i]$ to be equal
	to $min(D(X[i]))$. Therefore, $[L[1],\ldots, L[i]]$ is the lexicographically smallest prefix of length $i$ for a solution of $C(\myvec{X})$. 
	Hence, there is no solution of $C(\myvec{X})$ with prefix $[L'[1],\ldots,L'[i]] \leq_{lex} [L[1],\ldots,L[i]]$. This leads to a contradiction.
\end{enumerate}

\subsubsection{\label{ss:xlex_lb} A filtering algorithm for the $\LEXX_{lb}(\myvec{L},\myvec{X},C)$
constraint.}

{The propagation algorithm for the $\LEXX_{lb}(\myvec{L},\myvec{X},C)$
constraint finds all possible supports that are greater than or equal 
to the lower bound  $\myvec{L}$ and marks the values
that occur in these supports. 
Algorithm~\ref{a:lexx_rlx} gives the pseudo-code for the 
propagator for $\LEXX_{lb}$.
The algorithm uses 
the auxiliary routine $MarkConsistentValues(C,\myvec{X},\myvec{X'})$. 
This finds all values in domains of $\myvec{X'}$ 
that satisfy $C(\myvec{X'})$ and marks corresponding values in $\myvec{X}$. 
The time complexity of the $MarkConsistentValues(C,\myvec{X},\myvec{X'})$ procedure
is $O(nd + c)$.  The total time complexity of the propagator
for the $\LEXX_{lb}$ filtering algorithm is $O(n (nd +  c))$. 
A dual algorithm to $\LEXX_{lb}$ is $\LEXX_{ub} (\myvec{X},\myvec{U},C)$ that
finds all possible supports that are less than or equal 
to the upper bound $\myvec{U}$ and marks the values
that occur in these supports.

\begin{algorithm}
\scriptsize {
\caption{$\LEXX_{lb} (\myvec{L},\myvec{X},C)$ }\label{a:lexx_rlx}
\begin{algorithmic}[1]
\Procedure{$\LEXX_{lb}$}{$\myvec{L} : out ,\myvec{X} : out,C : in$}
\If {$(DC(C(\myvec{X})) == fail)$}\label{a:lexx_rlx_dc}
	\State return $false$; \label{a:lexx_rlx_fail} 
\EndIf
\State $\myvec{LX} = \myvec{X}$;
\For{$i = 1$ \textbf{to} $n$} \label{}	
			\State	$D(LX[i]) =  \{v_j|  v_j \in D(LX[i])\ and \ L[i] < v_j\}$;
			\State $MarkConsistentValue(C, \myvec{X}, \myvec{LX})$;\label{a:lexx_rlx_mark_2}
			\If {$L[i] \notin D(X[i])$}\label{a:lexx_rlx_check_2}
				\State break;		
			\Else	
				\State $LX[i] = L[i]$; 
			\EndIf		
			
\EndFor 
\If {($i==n$)}
			\State $MarkConsistentValues(C, \myvec{X}, \myvec{L})$;\label{a:lexx_rlx_mark_1}
\EndIf	
\For{$i = 1$ \textbf{to} $n$} \label{}			
	\State $Prune ( \{v_j \in D(X[i])| unmarked(v_j) \})$; 
\EndFor 
\EndProcedure
\end{algorithmic}
}
\end{algorithm}

\begin{algorithm}
\scriptsize {
\caption{Mark consistent values}\label{a:gench_marking}
\begin{algorithmic}[1]
\Procedure{$MarkConsistentValues$}{$C : in,\myvec{X} : out,\myvec{X'} : in$}
\State $\myvec{Z} = Copy (\myvec{X'})$;
\State $DC(C(\myvec{Z}))$;
\For{$i = 1$ \textbf{to} $n$} \label{}	
	\State $Mark\{v_j| v_j \in D(X[i]) \ and \  v_j \in D(Z[i])\}$;
\EndFor 
\EndProcedure
\end{algorithmic}
}
\end{algorithm}

{
We also need to prove that 
Algorithm~\ref{a:lexx_rlx}  enforces domain consistency on the $\LEXX_{lb}(\myvec{L},\myvec{X},C)$ constraint.}
A dual proof holds for $\LEXX_{ub}$. 
\begin {proposition}
\label{p:al_2_proof}
\nina{Algorithm~\ref{a:lexx_rlx} enforces $DC$ on the $\LEXX_{lb}(\myvec{L},\myvec{X},C)$ constraint.}
\end {proposition}
\proof
We first show that if a value $v$ was not pruned from the domain of $X[p]$ (or marked) then 
it does  have a support for $\LEXX_{lb} (\myvec{L},\myvec{X},C)$. 
We then show that 
if a value $v$ was  pruned from the domain of $X[p]$ (or not marked) then 
it does not have a support. 
\begin{enumerate}
	\item Algorithm~\ref{a:lexx_rlx} marks values in two lines \ref{a:lexx_rlx_mark_2} and \ref{a:lexx_rlx_mark_1}. 
	Suppose at step $i$ the algorithm  marks value $v \in D(X[p])$ at line \ref{a:lexx_rlx_mark_2}. At this point we have
	that $LX[k]=L[k]$, $k=1,\ldots,i-1$, $L[i] < LX[i]$. After enforcing $DC$ on $C(LX)$, the value	$v$ is
	left in the domain of $LX[p]$. Consequently, there exists a support for $X[p]=v$, starting with 
	$[L[1],\ldots,L[i-1], v',\ldots]$, $v' \in D(LX[i])$,  that is strictly greater than $L$. 
	Marking at line~\ref{a:lexx_rlx_mark_1} covers the case where $\myvec{L}$ is a solution of $C(\myvec{X})$.	
	\item By contradiction. Suppose that value $v \in D(X[p])$ was not marked by Algorithm~\ref{a:lexx_rlx} but it has a 
	support $\myvec{X'}$ such that $\myvec{L} \leq_{lex} \myvec{X'}$.  Let $i$ be the first position where 
	$L[i] < X[i]$ and $L[k] = X[k]$, $k=1,\ldots,i-1$. We consider three disjoint cases: 
	
	\begin{itemize}
		\item The case that no such $i$ exists. Then $\myvec{L}$ is a support for value $v \in D(X[p])$. Hence, value $v$ has to 
		be marked at line~\ref{a:lexx_rlx_mark_1}. This leads to a contradiction.		
		\item {The case that $i \leq n$ and $ p < i$}. Note that in this case $v$ equals $L[p]$.
		Consider Algorithm~\ref{a:lexx_rlx} at step $i$. At this point
		we have $L[k] = LX[k]$, $k=1,\ldots,i-1$.  After enforcing $DC$ on $C(LX)$ (line~\ref{a:lexx_rlx_mark_2}), 
		values $X'[k]$, $i=1,\ldots,n$ are left in the domain of $\myvec{LX}$, because $L[i] < X'[i]$,
		$L[k] = X'[k]$, $k=1,\ldots,i-1$. Hence, value $v \in X'[p]$ will be marked at line~\ref{a:lexx_rlx_mark_2}. 
		 This leads to a contradiction.     
		\item {The case that $i \leq n$ and $i \leq p$}. Consider Algorithm~\ref{a:lexx_rlx} at step $i$. 
		At this point we have $L[k] = LX[k]$, $k=1,\ldots,i-1$. Moreover, value $X'[i]$ has to be in the domain of $LX[i]$,
		because value $X'[i]$ is greater than $L[i]$ and is consistent with the partial assignment $[L[1],\ldots, L[i-1]]$. Domains 
		of variables $\myvec{LX}$ contain all values that have supports starting with $[L[1],\ldots, L[i-1]]$ 
		and are strictly greater than $\myvec{L}$. Consequently, they contain $X'[i]$, $i=1,\ldots,n$ and the algorithm marks $v$ at 				
		line~\ref{a:lexx_rlx_mark_2}.	This leads to a contradiction.     
	\end{itemize}	 
\end{enumerate}
\qed

\subsubsection{\label{ss:xlex} A filtering algorithm for the $\LEXX(\myvec{X},\myvec{Y},C)$.}
\ninaS{Algorithm~\ref{a:lexx} enforces domain consistency on the $\LEXX(\myvec{X},\myvec{Y}, C)$
constraint. Following Proposition~\ref{p:p1}, Algorithm~\ref{a:lexx} finds the lexicographically smallest (greatest) solutions for 
$C(\myvec{X})$ ($C(\myvec{Y})$) and runs a relaxed version of $\LEXX$ for each row.}
Algorithm~\ref{a:lexx} gives the pseudo-code for the 
propagator for the $\LEXX(\myvec{X},\myvec{Y},C)$ constraint.
}
\begin{algorithm}
\scriptsize {
\caption{$\LEXX (\myvec{X},\myvec{Y},C)$ }\label{a:lexx}
\begin{algorithmic}[1]
\Procedure{$\LEXX$}{$\myvec{X} : out,\myvec{Y} : out,C : in$}
\If {$(C_{min}(\myvec{X_l},\myvec{X}) == fail)$ or $(C_{max}(\myvec{Y},\myvec{Y_u})) == fail)$} \label{a:lexx_min_max}
	\State return $false$;
\EndIf
\If {($\myvec{X_l} >_{lex}  \myvec{Y_u}$)}
	\State return $false$; \label{a:lexx_fail}
\EndIf
\State $\LEXX_{lb}(\myvec{X_l},\myvec{Y},C)$; \label{a:lexx_Y}
\State $\LEXX_{ub}(\myvec{X},\myvec{Y_u},C)$; \label{a:lexx_X}
\EndProcedure
\end{algorithmic}
}
\end{algorithm}  

\begin {proposition}
\label{p:al_3_proof}
\nina{Algorithm~\ref{a:lexx} enforces $DC$ on the $\LEXX(\myvec{X},\myvec{Y}, C)$ constraint.}
\end {proposition}
\proof
Correctness of the algorithm follows from correctness of the decomposition (Proposition~\ref{p:p1}).
However, we need to consider the case where $\myvec{X_l} >_{lex} \myvec{Y_u}$,  prove 
correctness of the $\LEXX_{lb}$ and $\LEXX_{ub}$ algorithms and prove that the algorithm only needs to run once.

If $\myvec{X_l} >_{lex} \myvec{Y_u}$ then $\LEXX(\myvec{X},\myvec{Y},C)$ does not have  a solution and Algorithm~\ref{a:lexx}
fails at line~\ref{a:lexx_fail}. Otherwise, we notice that 
if $\myvec{X_l} \leq_{lex} \myvec{Y_u}$ then $\myvec{X_l}$ and $\myvec{Y_u}$ is a solution of 
$\LEXX(\myvec{X},\myvec{Y},C)$, because $\myvec{X_l}$ is a solution of $C(\myvec{X})$, $\myvec{Y_u}$ is a solution of $C(\myvec{Y})$
and  $\myvec{X_l} \leq_{lex} \myvec{Y_u}$. Consequently, invocation of the simplified version of $\LEXX$
at lines~\ref{a:lexx_Y} and~\ref{a:lexx_X} cannot change $\myvec{X_l}$ and $\myvec{Y_u}$.
\qed

\begin{example}
We consider how Algorithm~\ref{a:lexx} works on the first two rows $\LEXX$ constraint from
Example~\ref{e:sep}. Let $n$ equal $5$. In this case domains of 
the first two rows of variables are 
$\left(
\begin{matrix}	
  M[1] \\
  M[2] 
\end{matrix}    
\right)=\left(
\begin{matrix}	
  [1,2,3,4] & [6,7,8,9] & 5 \\
  [1,2,3,4] & [5,6,7,8] & 4\\
\end{matrix}    
\right)$.

Suppose the solver branches on $X[1] = 1$. Algorithm~\ref{a:lexx} finds
the lexicographically smallest and greatest solutions of $M[1]$ and $M[2]$
using Algorithm~\ref{a:c_min}(line~\ref{a:lexx_min_max}). These solutions
are $[1,6,5]$ and $[4,8,4]$ respectively . 
Then enforces $DC$ on $\LEXX_{lb} ([1,6,5],M[2],C)$ in the following way:
\begin{enumerate}
	\item copies $M[2]$ to $\myvec{LX}$
	\item marks all values that have a support starting with a value greater than $1$ (that is $2$, $3$ and $4$).
	There are three supports that satisfy this condition, namely, $[2,6,4]$, $[3,7,4]$ and $[4,8,4]$. Checks conditions
	at line~\ref{a:lexx_rlx_check_2} and assigns $LX[1]$ to $1$. Then it moves to the next iteration.
	\item marks all values that have a support starting with a prefix greater than $[1,6]$. There are no such values.
	 Checks conditions at line~\ref{a:lexx_rlx_check_2} and 
	 assigns $LX[2]$ to $6$. Then it moves to the next iteration.
	\item marks all values that have a support starting with a prefix greater than $[1,6,5]$. There are no such values.
	 Checks  conditions at line~\ref{a:lexx_rlx_check_2}  and stops the marking part.
	\item removes unmarked values: value $1$ from $X[2]$ and value $5$ from $Y[2]$.  
\end{enumerate}

Finally, it enforces $DC$ on $\LEXX_{ub} (M[1],[4,8,4],C)$. This sets $M[1]$ to $[1,6,5]$, because the solver
branched on  $X[1] = 1$ and $[1,6,5]$ is the only possible support for this assignment.  

\end{example}

The time complexity of the general algorithm is more expensive than the decomposition into 
individual constraints $C(\myvec{X})$, $C(\myvec{Y})$ and $\LEX(\myvec{X},\myvec{Y})$ by a linear factor. 
The general algorithm is not incremental, \ninaS {but its performance can be improved by detecting
entailment. If $\myvec{X_u} < \myvec{Y_l}$ then the $\LEX$ constraint is entailed and $\LEXX$ can 
be decomposed into two constraints $C(\myvec{X})$ and $C(\myvec{Y})$}. 
Similarly, we can improve the complexity by
detecting when $C(\myvec{X})$ and $C(\myvec{Y})$ are
entailed. As we show in the next 
sections, the time complexity of the
propagator for the $\LEXX(\myvec{X},\myvec{Y},C)$ constraint 
can also be improved by making it incremental for many common constraints $C$ 
by exploiting properties of $C$. Note also that Algorithm~\ref{a:lexx} easily extends
to the case that different global constraints are applied to $\myvec{X}$ and $\myvec{Y}$.  

%



\subsection {The $\LEXX(\myvec{X},\myvec{Y},\SEQUENCE)$ constraint}
\label{s:seq_lex}
{
In this section we consider the case of a conjunction of the $\LEX$ constraint with two $\SEQUENCE$ constraints.
First we assume that variables $\myvec{X}$ and $\myvec{Y}$ are Boolean variables. Later we will show how to extend this to the  general case.
In the Boolean case, we can exploit properties of the 
filtering algorithm for the $\SEQUENCE$ constraint ($\hprs$) proposed in ~\cite{Hoeve06}. 
The core of the $\hprs$ algorithms is the \verb|CheckConsistency| procedure 
that detects inconsistency if the $\SEQUENCE$ constraint is unsatisfiable and returns 
the lexicographically smallest solution otherwise. The $\hprs$ algorithm runs 
\verb|CheckConsistency| for each variable-value pair $X_i=v_j$.
If \verb|CheckConsistency| detects a failure, then value $v_j$ can be pruned from $D(X_i)$, otherwise
\verb|CheckConsistency| returns the lexicographically smallest support for $X_i=v_j$.    
As was shown in ~\cite{Hoeve06}, the algorithm can be modified
so that  \verb|CheckConsistency|  returns the lexicographically
greatest support.  
Both versions of the algorithm are useful for us. We will use the ${min}$ subscript for
 the first version of the algorithm, and the ${max}$ subscript for the second.

Due to these properties of the $\hprs$ algorithm, 
a propagator for the $\LEXX $ $(\myvec{X},\myvec{Y_u},\SEQUENCE)_{lb}$, denoted  $\hprs'_{min}(\myvec{X},\myvec{Y_u})$, is
a slight modification of $\hprs_{min}$, which checks that the lexicographically
smallest support for $X_i = v_j$ returned by  the $\verb|CheckConsistency|_{min}$ procedure is lexicographically
smaller than or equal to $\myvec{Y_u}$. To find the lexicographically greatest solution, $\myvec{Y_u}$, of the $\SEQUENCE(\myvec{Y})$ constraint, we run $\verb|CheckConsistency|_{max}$ on variables $Y$. Dual reasoning is applied to 
the $\LEXX(\myvec{X_l},\myvec{Y},\SEQUENCE)_{ub}$ constraint. Algorithms~\ref{a:lexx_seq} shows pseudo code for 
$DC$ propagator for the $\LEXX (\myvec{X},\myvec{Y},\SEQUENCE)$ constraint.

\begin{algorithm}
\scriptsize {
\caption{$\LEXX (\myvec{X},\myvec{Y},\SEQUENCE)$ }\label{a:lexx_seq}
\begin{algorithmic}[1]
\Procedure{$\LEXX$}{$\myvec{X} : out,\myvec{Y} : out, \SEQUENCE(l,u,k) : in$}
\If {$\neg(CheckConsistency_{min}(\myvec{X_l},\myvec{X}))$ or $\neg (CheckConsistency_{max}(\myvec{Y},\myvec{Y_u}))$} \label{a:lexx_seq_min_max}
	\State return $false$;
\EndIf
\If {($\myvec{X_l} >_{lex}  \myvec{Y_u}$)}
	\State return $false$; \label{a:lexx_seq_fail}
\EndIf
\State $\hprs'_{max}(\myvec{X_l},\myvec{Y},\SEQUENCE(l,u,k))$; \label{a:lexx_seq_Y}
\State $\hprs'_{min}(\myvec{X},\myvec{Y_u},\SEQUENCE(l,u,k))$; \label{a:lexx_seq_X}
\EndProcedure
\end{algorithmic}
}
\end{algorithm}  

$\hprs'_{min}$ and $\hprs'_{max}$ are incremental algorithms, therefore the total time
complexity of Algorithm ~\ref{a:lexx_seq} is equal to the complexity of the $\hprs$ algorithm, which is $O(n^3)$
down a branch of the search tree.  \ninaS{Correctness of Algorithm ~\ref{a:lexx_seq} follows from Proposition ~\ref{p:p1} and
correctness of the $\hprs$ algorithm.}

\begin{example} 
\label{e:seq}
\sloppy{
Consider the $\SEQUENCE(2,2,3,[X[1],X[2],X[3],X[4]])$ and 
$\SEQUENCE(2,2,3,[Y[1],Y[2],Y[3],Y[4]])$ constraints. 
The domains of the variables are $X = [\{0,1\},\{1\},\{0,1\},\{0,1\}]$ and
$Y = [\{0,1\},\{0,1\},\{1\},\{0,1\}]$.
Note that each of the two $\SEQUENCE$ and  the $\LEX (\myvec{X},\myvec{Y})$  constraints are domain consistent. 

The $\LEXX(\myvec{X},\myvec{Y},\SEQUENCE)$ constraint fixes  variables $\myvec{X}$ to $[0,1,1,0]$.
The lexicographically greatest solution for the $\SEQUENCE (Y)$ is $[1,0,1,0]$, while the 
lexicographically smallest support for $X[1]=1$ is $[1,1,0,1]$. Therefore, the value $1$ will
be pruned from the domain of $X[1]$. For the same reason, the value $0$ will be pruned from $X[3]$ and
the value $1$ will be pruned from $X[4]$.}
\end{example}

Consider the general case, where $\myvec{X}$ and $\myvec{Y}$ are finite domain variables.
We can channel the variables $\myvec{X}$, $\myvec{Y}$ into Boolean variables 
$\myvec{b_X}$,$\myvec{b_Y}$ and post $\SEQUENCE(\myvec{b_X})$, $\SEQUENCE(\myvec{b_Y})$, which does not hinder propagation.
Unfortunately, we cannot post  the $\LEX$ constraint on the Boolean variables $\myvec{b_X}$ and $\myvec{b_Y}$,
because some solutions will be lost.  For example, suppose we have
$\SEQUENCE(\myvec{X}, 0,1,2,\{2,3\})$ and $\SEQUENCE(\myvec{Y}, 0,1,2,\{2,3\})$ constraints. Let $\myvec{X} = [2,0,2]$ and $\myvec{Y} = [3,0,0]$ be solutions of these constraints. 
The corresponding Boolean variables are $\myvec{b_X} = [1,0,1]$ and $\myvec{b_Y}=[1,0,0]$.
Clearly $\myvec{X} <_{lex} \myvec{Y}$, but $\myvec{b_X} >_{lex} \myvec{b_Y}$. Therefore, the $\LEX$ constraint can be enforced only on the original variables.

The problem is that the $\hprs$ algorithm returns the lexicographically smallest solution on Boolean variables.
As the example above shows,  lexicographical comparison between Boolean solutions of $\SEQUENCE$s $\myvec{b_X}$ and $\myvec{b_Y}$ 
is not sound with respect to the original variables. Therefore, given a solution of $\SEQUENCE(\myvec{b_X})$, we need to find  
the corresponding lexicographically smallest solution of $\SEQUENCE(\myvec{X})$. We observe that if we restrict ourselves to a special case of  $\SEQUENCE (l,u,k,v,\myvec{X})$ where $\max (D\setminus v) < \min (v)$ then this problem can be solved in linear time as follows. Let $\myvec{b_X}$ be a solution for  
$\SEQUENCE(\myvec{b_X})$. Then the corresponding lexicographically smallest solution $\myvec{X}$ for 
$\SEQUENCE(\myvec{X})$ is $X[i] = \ min (v \cap D(X[i]))$ if $b_X[i] = 1$ and $X[i]= \ min (D(X[i]))$ otherwise. 
In a similar way we can find the corresponding lexicographically greatest solution.
A slight modification to Algorithm~\ref{a:lexx_seq} is needed in this case. 
Whenever we need to check whether $\myvec{b_X}$ is smaller than or equal to $\myvec{b_Y}$, we transform 
$\myvec{b_X}$ to the corresponding lexicographically smallest solution, $\myvec{b_Y}$   
to the corresponding lexicographically greatest solution and perform the comparison.         
}

\subsection {The $\LEXX(\myvec{X},\myvec{Y},\REGULAR)$ constraint}
\label{s:reg_lex}
\begin{figure}[htb] \centering
    \includegraphics[width=0.3\textwidth]{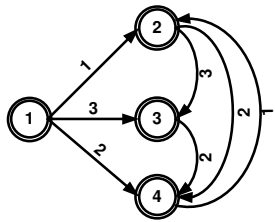}\\
    \caption{\label{f:f1} Automation for Example~\ref{e:reg}}
\end{figure}

With the $\REGULAR(\cal {A},\myvec{X})$ constraint, we 
will show that  we can build a propagator for \LEXX\ 
which takes just $O(nT)$ time, compared to $O(n^2T)$
for our general purpose propagator, where $d$ is
the maximum domain size and $T$ is the number of transitions of the automaton $\cal {A}$. 
We will use the following example to illustrate results in this section.
\begin{example}
\label{e:reg}
Consider the  $\LEXX(\myvec{X},\myvec{Y}, C)$ constraint where
the $C$ is $\REGULAR ({\cal {A}},\myvec{X})$ and $\cal {A}$ is the automaton presented in Figure \ref{f:f1}. 
Domains of variables are $X[1] \in \{1,2\}$, $X[2] \in \{1,3\}$, $X[3] \in \{2\}$
and $Y[1] \in \{1,2,3\}$, $Y[2] \in \{1,2\}$, $Y[3] \in \{1,3\}$.
\end{example}
Consider Algorithm~\ref{a:c_min} that finds the lexicographically smallest solution
of the $\REGULAR$ constraint. At line~\ref{a:s_c_min_dc} it invokes a $DC$
propagator for the $\REGULAR$ constraint to ensure that an
extension of a partial solution on each step leads to a 
solution of the constraint. To do so, it prunes 
all values that are inconsistent with the current partial
assignment.  We will show that for the $\REGULAR$ constraint
values {consistent} with the current partial assignment 
can be found in $O(log(d))$ time. 

\begin{figure}[htb] \centering
    \includegraphics[width=0.5\textwidth]{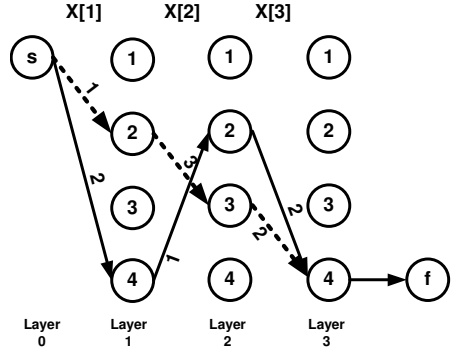}\\
    \caption{\label{f:f2} The $\REGULAR_{min}({\cal A}, \myvec{L},\myvec{X})$ algorithm. Dashed arcs correspond
    to the lexicographically smallest solution.}
\end{figure}

Let $G_x$ be a layered graph for the $\REGULAR$ constraint
and $\myvec{L_i}=[L[1],\ldots,L[i]]$ be a partial assignment 
at the $i$th iteration of the loop (lines \ref{a:s_reg_l_for} - \ref{a:e_reg_l_for}, Algorithm~\ref{a:reg_l}).
Then $\myvec{L_i}$ corresponds to a path from the initial node 
at $0$th layer to a node $q^i_j$ at $i$th layer. Clearly, values of 
$X[i+1]$ consistent with the partial assignment $\myvec{L_i}$ are labels of outgoing arcs from
the node $q^i_j$.  We can find the label with the minimal value in $O(log(d))$
time. Algorithm~\ref{a:reg_l} shows pseudo-code for $\REGULAR_{min}({\cal A}, \myvec{L},\myvec{X})$. 
Figure \ref{f:f2} shows a run of $\REGULAR_{min}({\cal A}, \myvec{L},\myvec{X})$ for variables $\myvec{X}$ in 
Example~\ref{e:reg}. \nina{The lexicographically smallest solution corresponds to dashed arcs}.

\begin{algorithm}
\scriptsize {
\caption{$\REGULAR_{min}({\cal A}, \myvec{L},\myvec{X})$ }\label{a:reg_l}
\begin{algorithmic}[1]
\Procedure{$\REGULAR_{min}$}{${\cal A} : in , \myvec{L} : out ,\myvec{X} : in$}
\State Build graph $G_x$;
\State $q[0] = q^0_0$;
\For{$i = 1$ \textbf{to} $n$} \label{a:s_reg_l_for}						
	\State $L[i] = \min \{v_j|v_j \in outgoing\_arcs(q[i-1]) \}$; \label{a:s_reg_check2}
	\State $q[i] = t_{{\cal {A}}}(q[i-1], L[i])$; \Comment {$t_{{\cal {A}}}$ is the transition function of ${\cal A}$.}
\EndFor \label{a:e_reg_l_for}			
\State return $L$;
\EndProcedure
\end{algorithmic}
}
\end{algorithm}

\begin{figure}[htb] \centering
\includegraphics[width=0.8\textwidth]{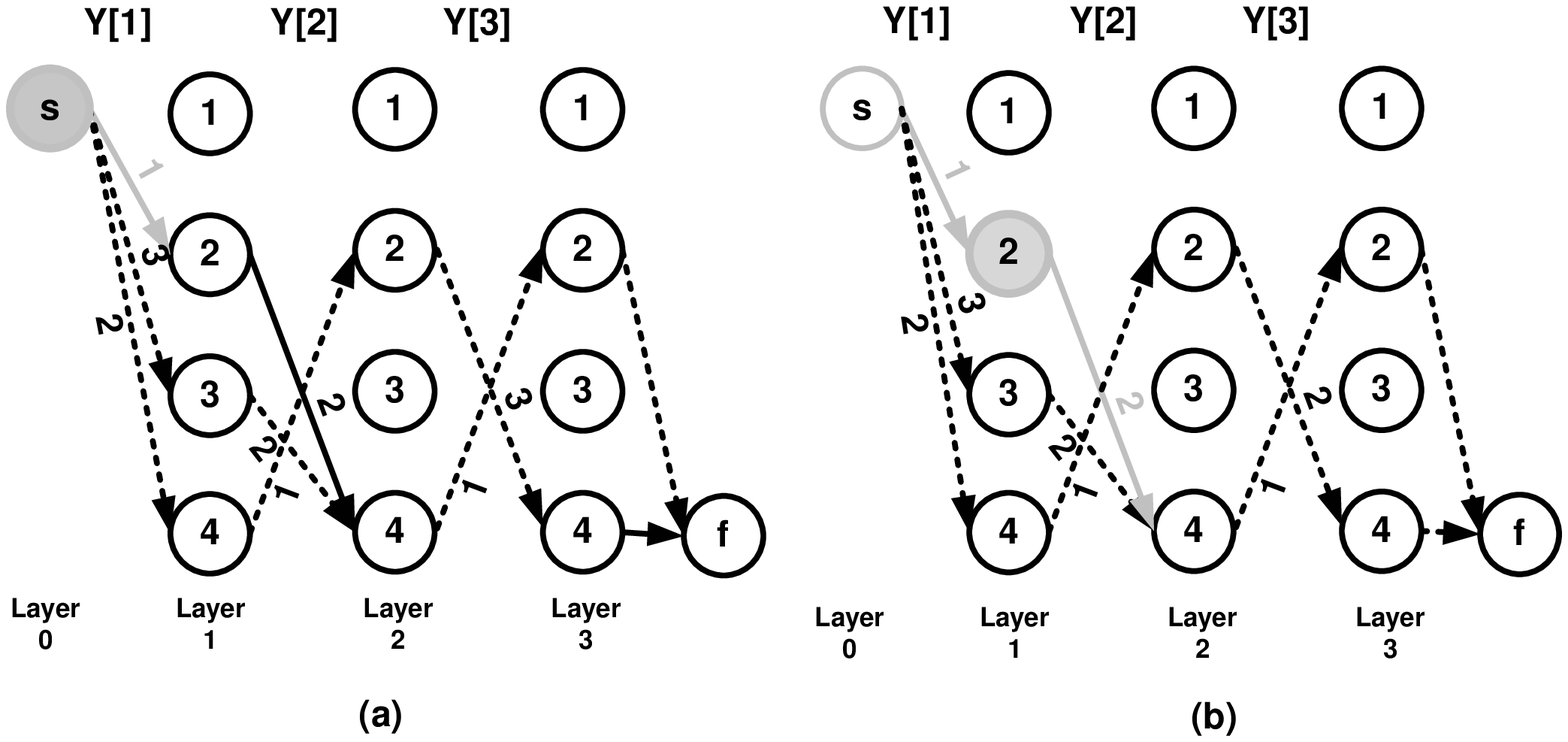} \\ 
\caption{\label{f:f3} A run of the $\LEXX_{lb} (\myvec{L},\myvec{X},\REGULAR)$ algorithm. Dashed arcs are marked.}
\end{figure}

The time complexity of Algorithm~\ref{a:lexx_rlx} for the $\REGULAR$ constraint is 
also $O(nT)$. The algorithm works with the layered graph rather than original variables.
On each step it marks edges that occur in feasible paths in $G_x$ that are lexicographically
greater than or equal to $\myvec{L}$. Figure \ref{f:f3} shows  execution of $\LEXX_{lb} (\myvec{X_l},\myvec{Y},\REGULAR)$ for variables $\myvec{Y}$ 
and the lexicographically smallest solution for $\myvec{X}$, $\myvec{X_l}=(1,3,2)$, from Example~\ref{e:reg}.  
It starts at initial node $s$ and marks all arcs on feasible paths starting with values greater than $X_l[1] = 1$ (that is $2$ or $3$).
\nina{Figure \ref{f:f3}(a)  shows the removed arc in gray and marked arcs in dashed style.}
Then, from the initial node at $0$th layer it moves to the $2$nd node at the $1$st layer (Figure \ref{f:f3} (b)). 
The algorithm marks all arcs on paths starting with a prefix greater than $[X_l[1], X_l[2]] =[1, 3]$. There are no such feasible paths.
So the $MarkConsistentArcs$ algorithm does not mark extra arcs. 
Finally, it finds that there is no outgoing arc from the $2$nd node at $2$nd layer labeled with $3$ and stops its marking 
phase. There are two unmarked arcs that are solid gray arcs at  Figure \ref{f:f3} (b).
The algorithm prunes value $1$ from the domain of $Y[1]$, because there are no marked arcs labeled with value $1$ for $Y[1]$.
Algorithm~\ref{a:reg_rlx} shows the pseudo-code for $\LEXX_{lb} (\myvec{L},\myvec{X},\REGULAR)$. 
Note that the $MarkConsistentArcs$ algorithm for the $\REGULAR$ constraint
is incremental. The algorithm performs a constant number of operations (deletion, marking) on each edge. Therefore,
the total time complexity is $O(nT)$ {at each invocation} of the $\LEXX_{lb} (\myvec{L},\myvec{X},\REGULAR)$  constraint.
\begin{algorithm}
\scriptsize {
\caption{Mark consistent arcs}\label{a:reg_marking}
\begin{algorithmic}[1]
\Procedure{$MarkConsistentArcs$}{$G_x : out, q : in$}
\State  Mark all arcs that occur on a path from $q$ to the final node;
\EndProcedure
\end{algorithmic}
}
\end{algorithm}

\begin{algorithm}
\scriptsize {
\caption{$\LEXX_{lb} (\myvec{L},\myvec{X},\REGULAR)$ }\label{a:reg_rlx}
\begin{algorithmic}[1]
\Procedure{$\LEXX_{lb}$}{$\myvec{L} : in,\myvec{X} : out ,\REGULAR : in $}
\State Build graph $G_x$;
\State $q[0] = q^0_0$;
\State $q_L = 0$;
\For{$i = 1$ \textbf{to} $n$} \label{}
			\State Remove outgoing arcs from the node $q[i-1]$ labeled with $\{min(X[i]),\ldots,L[i]\}$;
			\State $MarkConsistentArcs(G_x,q[i-1])$;\label{a:reg_rlx_del_lb}
			\If {$(\exists$ a outgoing arc from $q[i-1]$)$\wedge (i \neq 1)$}
				\State mark arcs $(q[k-1],q[k])$, $k=q_L,\ldots,i-1$ ;	
				\State $q_L=i-1$;
			\EndIf							
			\If {$L[i] \notin D(X[i])$}
				\State break;
			\EndIf
			\State $q[i] = T_{{\cal A}}(q[i-1], L[i])$; \Comment {$T_{{\cal A}}$ is the transition function of ${\cal A}$.}
\EndFor \label{}		
\If {($i==n$)}
			\State mark arcs $(q[k-1],q[k])$, $k=q_L,\ldots,n$; 	
\EndIf
	
\For{$i = 1$ \textbf{to} $n$} \label{}			
\State $Prune ( \{v_j \in D(X[i])| unmarked(v_j) \})$; 
\EndFor \label{}			
\EndProcedure
\end{algorithmic}
}
\end{algorithm}  

The second algorithm that we propose  represents the $\LEXX(\myvec{X},\myvec{Y},\REGULAR)$
as a single automaton that is the product of automata for two $\REGULAR$
constraints and an automaton for $\LEX$. First, we create individual automata for each
of three constraints. Let $Q$ be the number of states for each  $\REGULAR$
constraint and $d$ be the number of states for the $\LEX$ constraint. 
Second, we interleave 
the variables $\myvec{X}$ and $\myvec{Y}$, to get the sequence 
$X[1],Y[1],X[2],Y[2],\ldots, X[n],Y[n]$.  The resulting
automaton is a product of individual automata that works on the constructed sequence
of interleaved variables. The number of states of the
final automaton is $Q'=O(dQ^2)$. The total time complexity
to enforce $DC$ on the $\LEXX(\myvec{X},\myvec{Y},\REGULAR)$ constraint is thus
$O(nT')$, where $T'$ is the number of transitions of the product automaton. 
\ninaS{It should be noted that this algorithm is very easy to implement. Once the product 
automaton is constructed, we encode the $\REGULAR$ constraint for it as a set of ternary transition constraints~\cite{qwcp06}.
}

The third way to propagate the $\LEXX(\myvec{X},\myvec{Y},\REGULAR)$ constraint
is to encode it as a cost $\REGULAR$ constraint. 
W.L.O.G., we assume that there exist only one initial and one final state.
Let $G_x$ be the layered graph for 
$\REGULAR(\myvec{X})$ and $G_y$ be the layered graph for $\REGULAR(\myvec{Y})$.  We replace the final state at $n+1$th layer in
$G_x$ with the initial state at $0$th layer at $G_y$. Finally, we need to encode  $\LEX(\myvec{X},\myvec{Y})$ using the 
layered graph. We recall that the $\LEX(\myvec{X},\myvec{Y})$ constraint
can be encoded as an arithmetic constraint 
$(d^{n-1}X[1]+\ldots+d^0X[n] \leq d^{n-1}Y[1]+\ldots+d^0Y[n])$
or    
$(d^{n-1}X[1]+\ldots+d^0X[n] - d^{n-1}Y[1]-\ldots-d^0Y[n] \leq 0),$ where $d = |\bigcup_{i=1}^nD(X[i])|$.
 
\begin{figure}[htb] \centering
    \includegraphics[width=0.8\textwidth]{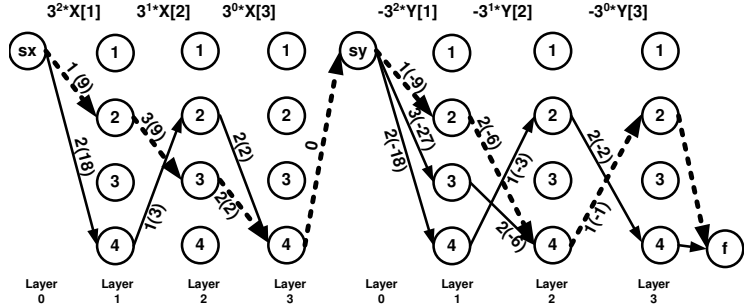}\\
    \caption{\label{f:f4} The $\LEXX (\myvec{X},\myvec{Y},\REGULAR)$ algorithm. Dashed arcs show the shortest path through the arc $(sy,2)$.}
\end{figure}

In turn this arithmetic constraint can be encoded in the layered graph by adding weights on corresponding arcs.
The construction for Example~\ref{e:reg} is presented in Figure \ref{f:f4}.  
Values in brackets are weights to encode the $\LEX(\myvec{X},\myvec{Y})$ constraint.
For instance, the arc $(\myvec{s}x,2)$ has weight $9$. The arc corresponds to the first variable with the
coefficient $d^2$, $d=3$. It is labeled with value $1$. The weight equals 
$1 \times d^2 =9$. More generally, an arc between the $k-1^{th}$ and $k^{th}$ layers 
labeled with $v_j$ is given weight $v_jd^{n-k}$.
Note that the weights of arcs that correspond to variables $\myvec{Y}$
are negative. Hence, the $\LEXX(\myvec{X},\myvec{Y},\REGULAR)$ constraint can be encoded as a cost 
$\REGULAR([{\cal A},{\cal A}],[\myvec{X},\myvec{Y}],W)$ constraint,
where $W$ is  the cost variable, $[{\cal A},{\cal A}]$ are two consecutive automata.  $W$ has to be less than or equal to $0$. 
Consider for example the shortest path through the arc $(sy,2)$. 
The cost of the shortest path through this arc is $3$. Consequently, value $1$
can be pruned form the domain of $Y[1]$.

The time complexity of enforcing $DC$ on the cost $\REGULAR ([{\cal A},{\cal A}], [\myvec{X }\myvec{Y}], W)$ is $O(nT)$,
{where $d=|\bigcup_{i=1}^nD(X[i])|$ and $T$ is the number of transitions of ${\cal {A}}$.} 
\footnote{Note that we have negative weights on arcs. However, we can add a constant $d^n$ to the weight of each arc 
and increase the upper bound of $W$ by this constant.} 
Again, the use of large integers adds a linear factor 
to the complexity, so we get $O(n^2T)$.



\section{Experimental results}
\sloppy{
To evaluate the performance of the proposed algorithms we carried out a series of experiments 
 nurse scheduling problems (NSP) for $\LEXX(\myvec{X},\myvec{Y},\SEQUENCE)$
and $\LEXX(\myvec{X},\myvec{Y},\REGULAR)$ constraints.  We used Ilog 6.2 for our experiments and
ran them on an Intel(R) Xeon(R)  E5405 $2.0$Ghz with $4$Gb of RAM. 
All benchmarks are modeled using a matrix model of $n \times m$ variables, 
where $m$ is the number of columns and $n$ is the number of rows.
}

\textbf {The $\LEXX(\myvec{X},\myvec{Y},\SEQUENCE)$ constraint.}
The instances for this problem are taken from www.projectmanagement.ugent.be/nsp.php. For
each day in the scheduling period, a nurse is assigned to a day, evening, or night shift
or takes a day off. The original benchmarks specify minimal required staff allocation
for each shift and individual preferences for each nurse. We ignore these preferences
and replace them with a set of constraints that model common workload restrictions for
all nurses. \ninaS{Therefore we use only labor demand requirements from the original benchmarks.}
We also convert these problems to Boolean problems by ignoring different shifts and only distinguishing whether 
the nurse  does or  does not work on the given day.
The labor demand for each day is the sum of labor demands for all shifts
during this day. In addition to the labor demand we post a single $\SEQUENCE$ constraint for each row.
\ninaS{We use a static variable ordering that assigns all columns in turn starting from the last one.} Each column is assigned from the bottom to the top. This tests if propagation can overcome a poor branching heuristic which conflicts 
with symmetry breaking constraints. 
We used six models with different  
$\SEQUENCE$ constraints posed on  rows of the matrix. 
\ninaS{Each model was run on 100 instances over a $28$-day scheduling period with $30$
nurses.}  Results are presented in Table~\ref{t:t2}. 
We compare $\LEXX(\myvec{X},\myvec{Y},\SEQUENCE)$  with the decomposition
into two $\SEQUENCE$ constraints and $\LEX$. In the case of the decomposition we used two algorithms
to propagate the $\SEQUENCE$ constraint. The first is the decomposition of the $\SEQUENCE$ constraint 
into individual $\AMONG$ constraints ($\among$), the second is the original $\hprs$ filtering
algorithm for  $\SEQUENCE$~\footnote{We would like to thank Willem-Jan van Hoeve for providing us with the implementation of the $\hprs$ algorithm.}. The decompositions are faster on easy instances 
that have a small number of backtracks, while they can not solve harder instances within the time limit. 
Overall, the model with the $\LEXX(\myvec{X},\myvec{Y},\SEQUENCE)$ constraint performs about 
$4$ times fewer backtracks and solves about $80$ more instances compared to the decompositions.

\begin{table}
\begin{center}
{\small
\caption{\label{t:t2} Simplified NSPs. 
Number of instances solved in 60 sec / average time to solve.
}
\begin{tabular}{|  c c |cc|cc|cc| cc|cc|}
\hline
$$ & $$
&\multicolumn {2}{|c|}{$\among$, $\LEX$}
&\multicolumn {2}{|c|}{$\hprs$, $\LEX$}
&\multicolumn {2}{|c|}{$\LEXX$} \\
\hline 
\hline 
 {1} &   \SEQUENCE(3,4,5)  & 46 & /   1.27 & 46 & /   2.76 & \textbf{74}& / \textbf{   1.44} \\ 
 {2} &   \SEQUENCE(2,3,4)  & 66 & /   0.63 & 66 & /   1.29 & \textbf{83}& / \textbf{   2.66} \\ 
 {3} &   \SEQUENCE(1,2,3)  & 20 & /   0.54 & 20 & /   1.04 & \textbf{34}& / \textbf{   3.17} \\ 
 {4} &   \SEQUENCE(4,5,7)  & 78 & /   1.36 & 77 & /   2.31 & \textbf{82}& / \textbf{   2.43} \\ 
 {5} &   \SEQUENCE(3,4,7)  & 55 & /   0.55 & 55 & /   1.07 & \textbf{58}& / \textbf{   1.53} \\ 
 {6} &   \SEQUENCE(2,3,5)  & 19 & /   5.38 & 18 & /   8.27 & \textbf{31}& / \textbf{   1.74} \\ 
\hline 
\hline 
\multicolumn {2}{|r|} {solved/total}& 284 &/600& 282 &/600& \textbf{362}& /600\\ 
\multicolumn {2}{|r|} {avg time for solved}& \multicolumn {2}{|c|}{  1.230} & \multicolumn {2}{|c|}{  2.194} & \multicolumn {2}{|c|}{\textbf{  2.147}} \\ 
\multicolumn {2}{|r|} {avg bt for solved}& \multicolumn {2}{|c|}{  18732} & \multicolumn {2}{|c|}{  16048} &\multicolumn {2}{|c|}{\textbf{   4382}} \\ 

\hline 
\end{tabular}}
\end{center}
\end{table}

\begin{table}
\begin{center}
{\small
\caption{\label{t:t3} NSPs.
Number of instances solved in 60 sec / average time to solve.
}
\begin{tabular}{|  c  |cc|cc|}
\hline
 
&\multicolumn {2}{|c|}{$\REGULAR$, $\LEX$}
&\multicolumn {2}{|c|}{$\LEXX$}\\
\hline 
\hline 
12 hours break                         & 30 & /   9.31 & \textbf{93}& / \textbf{   2.59} \\ 
 12 hours break + 2 consecutive shifts  & 87 & /   1.05 & \textbf{88}& / \textbf{   0.22} \\ 
\hline 
\hline 
\multicolumn {1}{|r|} {solved/total}& 117 &/200& \textbf{181}& /200\\ 
\multicolumn {1}{|r|} {avg time for solved}& \multicolumn {2}{|c|}{  3.166} & \multicolumn {2}{|c|}{\textbf{  1.439}} \\ 
\multicolumn {1}{|r|} {avg bt for solved}& \multicolumn {2}{|c|}{  35434} &\multicolumn {2}{|c|}{\textbf{   1220}} \\ 
\hline 
\end{tabular}}
\end{center}
\end{table}

\textbf {The $\LEXX(\myvec{X},\myvec{Y},\REGULAR)$ constraint.}
We implemented the second algorithm from Section~\ref{s:reg_lex}, which
propagates $\LEXX(\myvec{X},\myvec{Y},\REGULAR)$ using 
a product of automata for two $\REGULAR$ constraints
and the automaton for the $\LEX$ constraint. 
$\LEXX(\myvec{X},\myvec{Y},\REGULAR)$ was compared with  
decomposition into individual $\REGULAR$ and $\LEX$ constraints.
We used two models with different 
$\REGULAR$ constraints posed on  rows of the matrix. 
\ninaS{Each model was run on $100$ instances over a $7$-day scheduling period with $25$
nurses. We use the same variable ordering as above.} 
The $\REGULAR$ constraint in the first model expresses that 
each nurse should have at least 12 hours of break
between 2 shifts.
The $\REGULAR$ constraint in the second model 
expresses that each nurse should have at least 12 hours of break
between 2 shifts and at least two consecutive days on any shift.
Results are presented in Table~\ref{t:t3}. 
The model with the $\LEXX(\myvec{X},\myvec{Y},\REGULAR)$ constraint      
solves $64$ more instances than  decompositions and shows better run times and takes fewer backtracks.

\section{Related and future work}

Symmetry breaking constraints have on the whole
been considered separately to problem constraints.
The only exception to this of which we are aware is
a combination of lexicographical ordering and sum
constraints \cite{hkwaimath04}. This demonstrated
that on more difficult problems, or when the
branching heuristic conflicted with the symmetry
breaking, the extra pruning provided by the
interaction of problem and symmetry breaking
constraints is worthwhile. 
\ninaS{ Our work supports these results. Experimental results show that
using a combination of $\LEX$ and other global constraints achieves 
significant improvement in the number of backtracks and run time. Our future work is
to construct a filtering algorithm for the conjunction of the Hamming distance constraint with 
other global constraints. This is useful for modeling scheduling problems where we would like
to provide similar or different schedules for employees. We expect that performance improvement will be even greater than for
the $\LEXX$ constraint, because the Hamming distance constraint is much tighter than the $\LEX$ constraint.}


\bibliographystyle{splncs}


\bibliography{lit}

\begin{thebibliography}{10}

\bibitem{puget93}
Puget, J.F.:
\newblock On the satisfiability of symmetrical constrained satisfaction
  problems.
\newblock In: Proceedings of ISMIS'93. (1993)  350--361

\bibitem{Walsh07Sym2}
Walsh, T.:
\newblock Breaking value symmetry.
\newblock In: Proc. of the 13th Int. Conf. on Principles
  and Practice of Constraint Programming, CP 2007. (2007)  880--888

\bibitem{matrix2002}
Flener, P., Frisch, A.M., Kzlltan, B.H.Z., Miguel, I., Walsh, T.:
\newblock Matrix modelling: Exploiting common patterns in constraint
  programming.
\newblock In: Proc. of the Int. Workshop on Reformulating
  Constraint Satisfaction Problems. (2002)  27--41

\bibitem{ffhkmpwcp2002}
Flener, P., Frisch, A., Hnich, B., Kiziltan, Z., Miguel, I., Pearson, J.,
  Walsh., T.:
\newblock Breaking row and column symmetries in matrix models.
\newblock In: Proc. of 8th Int. Conf. on Principles and
  Practice of Constraint Programming. (2002)  462--476

\bibitem{Frisch02}
Frisch, A., Hnich, B., Kiziltan, Z., Miguel, I., Walsh, T.:
\newblock Global constraints for lexicographic orderings.
\newblock In: Proc. of the 8th Int. Conf. on Principles and
  Practice of Constraint Programming (CP'02), van Hentenryck, P (2002)  93--108


\bibitem{Beldiceanu94CHIP}
Beldiceanu, N., Contejean, E.:
\newblock Introducing global constraints in {CHIP}.
\newblock Mathematical and Computer Modelling \textbf{12} (1994)  97--123

\bibitem{Pesant04}
Pesant, G.:
\newblock A regular language membership constraint for finite sequences of
  variables.
\newblock In: Proc. of 10th Int. Conf. on
  Principles and Practice of Constraint Programming (CP'04). (2004)  482--495

\bibitem{Carlsson02LEXCHAIN}
Carlsson, M., Beldiceanu, N.:
\newblock Arc-consistency for a chain of lexicographic ordering constraints.
\newblock TR T--2002-18, Swedish Institute of Computer Science
  (2002)

\bibitem{Hoeve06}
Hoeve, W.J.v., Pesant, G., Rousseau, L.M., Sabharwal, A.:
\newblock Revisiting the {S}equence {C}onstraint.
\newblock In: Proc. of the 12th Int.
  Conf. on Principles and Practice of Constraint Programming (CP '06). (2006)  620--634

\bibitem{qwcp06}
Quimper, C.G., Walsh, T.:
\newblock Global {G}rammar constraints.
\newblock In : Proc. of the 12th Int.
  Conf. on Principles and Practice of Constraint Programming.  (2006)  751--755

\bibitem{hkwaimath04}
Hnich, B., Kiziltan, Z., Walsh, T.:
\newblock Combining symmetry breaking with other constraints: lexicographic
  ordering with sums.
\newblock In: Proc. of the 8th Int. Sym. on the Artificial
  Intelligence and Mathematics. (2004)

\end{thebibliography}

\end{document}